\documentclass[twoside]{article}
\usepackage[accepted]{aistats2019}

\usepackage[vlined]{algorithm2e}
\usepackage{amsmath}
\usepackage{amsfonts}
\usepackage{amsthm}
\usepackage{comment}
\usepackage[T1]{fontenc}
\usepackage[utf8]{inputenc}

\usepackage{graphicx}
\usepackage{hyperref}
\usepackage{microtype}
\usepackage{nicefrac}
\usepackage{rotating}
\usepackage{subcaption}
\usepackage{url}

\usepackage{booktabs}
\usepackage{xcolor}

\usepackage[titletoc,toc,title]{appendix}

\usepackage[inline]{enumitem}
\usepackage{mathtools}

\newlist{inlinelist}{enumerate*}{1}
\setlist*[inlinelist,1]{%
  label=(\arabic*),
}

\newcommand{\bX}{\mathbf{X}}

\newcommand{\bR}{\mathbf{R}}
\newcommand{\E}{\mathbb{E}}
\usepackage{Definitions}
\theoremstyle{plain}
\newtheorem{assumption}{Assumption}
\newtheorem{lemma}{Lemma}

\begin{document}

\makeatletter
\def\@copyrightspace{\relax}
\makeatother

\twocolumn[

\aistatstitle{Local Orthogonal Decomposition for Maximum Inner Product Search}
\aistatsauthor{Xiang Wu \And Ruiqi Guo \And Sanjiv Kumar \And  David Simcha}
\aistatsaddress{Google Research, New York}

]

\begin{abstract}

Inverted file and asymmetric distance computation (IVFADC) have been successfully applied to approximate nearest neighbor search and subsequently maximum inner product search.
In such a framework, vector quantization is used for coarse partitioning while product quantization is used for quantizing residuals.
In the original IVFADC as well as all of its variants, after residuals are computed, the second production quantization step is completely independent of the first vector quantization step.
In this work, we seek to exploit the connection between these two steps when we perform non-exhaustive search.
More specifically, we decompose a residual vector locally into two orthogonal components and perform uniform quantization and multiscale quantization to each component respectively.
The proposed method, called local orthogonal decomposition, combined with multiscale quantization consistently achieves higher recall than previous methods under the same bitrates.
We conduct comprehensive experiments on large scale datasets as well as detailed ablation tests, demonstrating effectiveness of our method.

\end{abstract}

\section{Introduction}

Maximum inner product search (MIPS) has become a popular paradigm for solving large scale classification and retrieval tasks. For example, in recommendation systems, user queries and documents are embedded into dense vector space of the same dimensionality and MIPS is used to find the most relevant documents given a user query~\cite{MIPSRecSys}.
Similarly in extreme classification tasks~\cite{MIPSForEC}, MIPS is used to predict the class label when a large number of classes are involved, often on the order of millions or even billions.
Lately it has also been applied to training tasks such as scalable gradient computation in large output spaces~\cite{LossDecompHsu}, efficient sampling for speeding up softmax computation~\cite{MIPSSampledSoftmax} and sparse update in end-to-end trainable memory systems~\cite{NeuralEpisodicControl}.

Formally, MIPS solves the following problem. Given a database of vectors $\bX=\{ x_i \}_{[N]}$ and a query vector $q$, where both $x_i, \, q \in \bR^d$, we want to find $x^*_q \in \bX$  such that $x^*_q = \argmax_{x \in \bX} q \cdot x$.

Although related, MIPS is different from $\ell_2$ nearest neighbor search in that inner product (IP) is not a metric, and triangle inequality does not apply. We discuss this more in Section~\ref{sec:related}.

\subsection{Background}

We refer to several quantization techniques in this work and we briefly introduce their notations:
\begin{itemize}
    \item \textbf{Scalar Quantization (SQ)}: The codebook of SQ $B_{SQ} = \{y_i\}_{[n_{SQ}]}$ contains $n_{SQ}$ scalars. A scalar $z$ is quantized into $\phi_{SQ}(z) = \argmin_{y \in B_{SQ}} |z - y|$. The bitrate per input is $l_{SQ} = \lceil \log_2 n_{SQ} \rceil$.
    
    \item \textbf{Uniform Quantization (UQ)}: UQ is a specialization of SQ, whose codebook is parameterized with only 2 scalars: $B_{UQ} = \{ai + b\}_{[n_{UQ}]}$. Though the UQ codebook is restricted to this structure, its major advantage over SQ is that the codebook can be compactly represented with only 2 scalars.
    
    \item \textbf{Vector Quantization (VQ)}: VQ is a natural extension of scalar quantization into vector spaces. Give a codebook $C = \{ c_i \}_{[m]}$ with $m$ codewords, an input vector $x$ is quantized into: $\phi_{VQ}(x) =  \argmin_{c \in C} \|x - c\|_2$. And the code that we store for vector $x$ is the index of the closest codeword in the VQ codebook: $\textrm{index}_{VQ}(x)$.
    
    \item \textbf{Product Quantization (PQ)}: To apply PQ, we first divide a vector into $n_B$ subspaces: $x = x^{(1)} \oplus x^{(2)} \oplus \cdots \oplus x^{(n_B)}$. And within each subspace we apply an independent VQ with $n_W$ codewords, i.e., $\phi_{PQ}(x) = \oplus_{i \in [n_B]} \phi^{(i)}_{VQ}(x^{(i)})$. The bitrate per input for PQ is thus $n_B \lceil \log_2 n_W \rceil$.
\end{itemize}

The IVFADC~\cite{PQ} framework combines VQ for coarse partitioning and PQ for residual quantization:

\begin{itemize}
\item \textbf{IVF}: An inverted file is generated via a VQ partitioning. Each
VQ partition $P_i$ contains all database vectors $x$ whose closest VQ center is $c_i$, i.e., $P_i = \{x \in \bX| c_i = \argmin_{c \in C} \|x - c\|_2 \}$.
    Within each partition $P_i$, residual vectors $\{r_x = x - c_i\}_{ x \in P_i}$ are further quantized with PQ and we denote the quantized approximation of the residual $r_x$ as $\phi_{PQ}(r_x)$.
    \item \textbf{ADC}: Asymmetric distance computation refers to an efficient table lookup algorithm that computes the approximate IP. For VQ, $ADC(q, \phi_{VQ}(x)) = \textrm{lookup}(\{q \cdot c\}_{c \in C}, \textrm{index}_{VQ}(x))$. And for PQ with multiple subspaces, we can decompose the dot product as:
    \begin{eqnarray*}
    & q \cdot r_x \approx q \cdot \phi_{PQ}(r_x) = ADC(q, \phi_{PQ}(r_x)) = & \\
    & \sum_{i \in [n_B]} ADC^{(i)}(q^{(i)}, \phi^{(i)}_{VQ}(r_x^{(i)})) &
    \end{eqnarray*}

    \item \textbf{Non-Exhaustive Search}: When processing a query $q$, we use IVF to determine the top partitions according to $q \cdot c_i$. We select top $m_{ADC}$ partitions to search into and then apply ADC to residuals in these top partitions.
\end{itemize}

There are many variations of the IVFADC setup. For example, the codebooks of VQ partitioning and PQ quantization can be (jointly) learned, and asymmetric distance computation can be implemented with SIMD instructions~\cite{MSQ, BlalockBOLT}. We discuss these variations in depth and the relation to this work in the Section~\ref{sec:related}.

In large scale applications, as the database size increases, larger $m$ and $m_{ADC}$ are generally used in IVF. Auvolat et~al. in \cite{KMeansForMIPS} proposes to use $m \sim N^{1/2}$ for 1-level VQ partitions and $m \sim N^{1/3}$ for 2-level etc.
From latest publications~\cite{Deep10m, FAISS}, the number of partitions for large datasets is among $10^3 - 10^6$. Hence in the following discussion, we focus on the case where the number of partitions is much larger than the vector dimension, i.e., $m \gg d$.

The scale of modern MIPS systems is often limited by the cost of storing the quantized vectors in main memory.
Therefore, we focus on methods that operate under low bitrate and can still achieve high recall.
This is reflected in our experiments in Section~\ref{sec:experiments}.

\subsection{Empirical Study of Inner Product Variance}

The overall quality of IP approximation is crucially dependent on the \emph{joint} distribution of the query $q$ and the residual $r_x = x - c_i$, where $c_i$ is the center of the partition $P_i$ that $x$ is assigned to.
In the non-exhaustive setup, the fact that we search into partition $P_i$ reveals strong information about the local conditional query distribution.
Nonetheless, previous methods approximate $q \cdot r_x$ by first quantizing $r_x$ independent of $q$ distribution.
And a close analysis of the IP $q \cdot r_x$ clearly shows that its variance is distributed \emph{non-uniformly} in different directions.
Formally a direction is a unit norm vector $\|v\|_2 = 1$, and the the \emph{projected IP} on direction $v$ is defined as: $(q \cdot v)(r_x \cdot v)$.
Within a partition $P$, we define the projected IP variance along $v$ as $\textrm{Var}(v) = \frac{1}{|P|} \sum_{x \in P} ((q \cdot v)(r_x \cdot v))^2$.
Note that the empirical first moment $\frac{1}{|P|} \sum_{x \in P} (q \cdot u)(r_x \cdot u) = 0$ by construction of VQ partitions.

\begin{figure}[ht]
\begin{subfigure}[h]{0.48\linewidth}
\includegraphics[width=\linewidth]{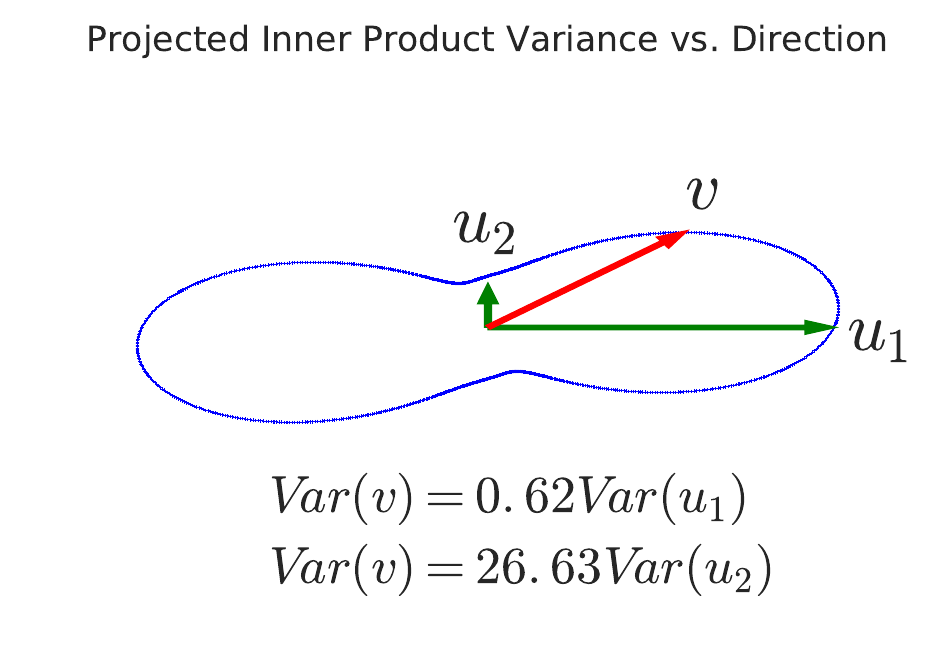}
\caption{}
\label{fig:var_vs_direction}
\end{subfigure}
\hfill
\begin{subfigure}[h]{0.48\linewidth}
\includegraphics[width=\linewidth]{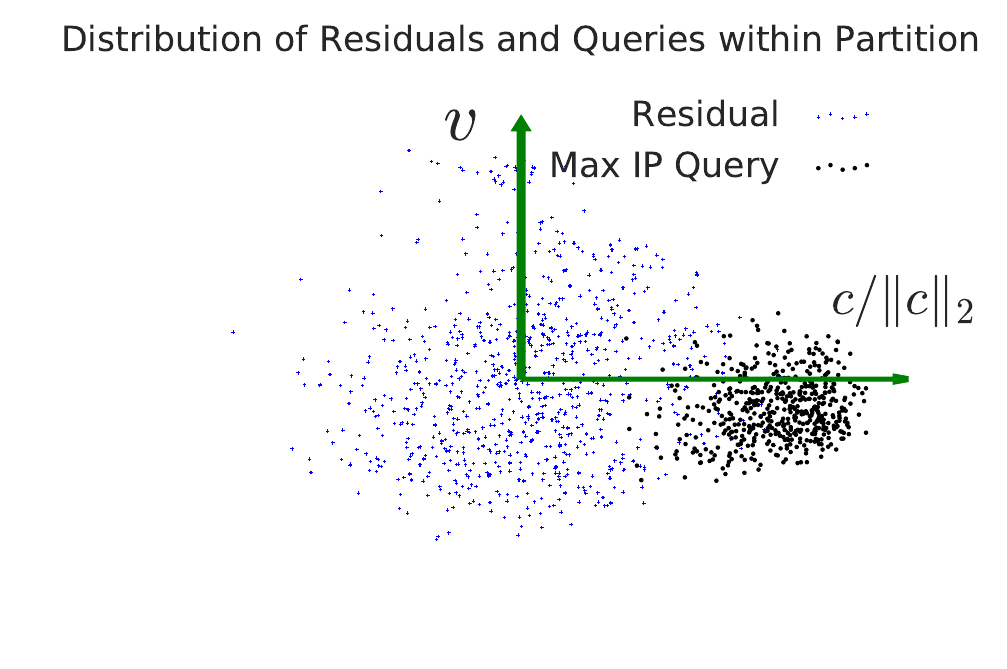}
\caption{}
\label{fig:rq_dist}
\end{subfigure}%
\caption{Non-uniform distribution of projected IP variance: (a) projected IP variance vs. angle in 2-dimensional subspace spanned by $\{u_1, u_2\}$. The projected IP variance is represented by the distance from the origin to the corresponding blue point at the angle. Variances are linearly scaled so that they fit the aspect ratio. (b) Scatter plot of residuals and queries that have maximum IPs with the partition center.}
\end{figure}

We conduct two different analyses with the public Netflix~\cite{BennettNetflixPrize} dataset.
In Figure~\ref{fig:var_vs_direction}, we fix the query $q$ and thus its top partition $P^*_q$ and its center $c^*_q$.
We pick the first direction $u_1 = c^*_q / \|c^*_q\|_2$ and the second direction $u_2$ orthogonal to $u_1$ randomly.
We then generate $n_v=1000$ evenly spaced directions in the subspace spanned by $\{ u_1, u_2 \}$ as: $\{v_i = \cos{(2i\pi / n_v)} u_1 + \sin{(2i\pi / n_v)} u_2 \}_{[n_v]}$.
We finally plot of the set of points $\{ ( \textrm{Var}(v_i)\cos{(2i\pi / n_v))}, \textrm{Var}(v_i)\sin{(2i\pi / n_v))}\}_{[n_v]}$, i.e., the distance between each point and the origin represents the projected IP variance on its direction.
The elongated peanut shape demonstrates clearly that variance of projected IPs is more concentrated on some directions than others.

In Figure~\ref{fig:rq_dist}, we fix a partition and plot 1) the residuals in the partition and 2) queries that have maximum IPs with the partition center.
We project all residuals and queries with maximum IPs onto the 2-dimensional subspace spanned by the partition center direction and the first principal direction of the residuals.
Residuals in blue are scattered uniformly in this subspace, but queries in black are much more concentrated along the direction of partition center $c/\|c\|_2$.

\subsection{Contributions}

This paper makes following main contributions:
\begin{itemize}
    \item Introduces a novel quantization scheme that directly takes advantage of the non-uniform distribution of the variance of projected IPs.
    \item Identifies the optimal direction for projection within each partition and proposes an effective approximation both theoretically and empirically.
    \item Designs complete indexing and search algorithms that achieve higher recall than existing techniques on widely tested public datasets.
\end{itemize}

\section{Related Work}
\label{sec:related}
The MIPS problem is closely related to the $\ell_2$ nearest neighbor search problem as there are multiple ways to transform MIPS into equivalent instances of $\ell_2$ nearest neighbor search.
For example, Shrivastava and Li~\cite{ALSH} proposed augmenting the original vector with a few dimensions. Neyshabur and Srebro proposed another simpler transformation to augment just one dimension to original vector: $\hat{x} = [ x / U; \sqrt{1 - (\|x\|_2 / U)^2}], \, \hat{q} = [q / \|q\|_2; 0]$. Empirically, the augmentation strategies do not perform strongly against
strategies that work in the unaugmented space.

\textbf{Learning Rotation and Codebooks}. Learning based variations of IVFADC framework have been proposed. One of the focuses is learning a rotation matrix which is applied before  vectors are quantized.
Such rotation reduces intra-subspace statistical dependence as analyzed in OPQ \cite{OPQ, CKmeans} and its variant \cite{LOPQ} and thus lead to smaller quantization error.
Another focus is learning codebooks that are additive such as~\cite{AQ, TreeAQ, AQRevisited, CQ, SCQ, SQ}. In these works, codewords are learned in the full vector space instead of a subspace, and thus are more expressive. Empirically, such additive codebooks perform better than OPQ at low bitrates but the gain diminishes at higher bitrates. 

\textbf{ADC Implementation}. ADC transforms inner product computations into a lookup table based operation, which can be implemented in different ways. The original ADC paper~\cite{PQ} used L1 cache based lookup table. Johnson et al.~\cite{FAISS} used an GPU implementation for ADC lookup. A SIMD based approach was also developed by~\cite{SIMDLUT, MSQ}. Again, this is orthogonal to the local decomposition idea of this work, as any ADC implementation can be used in this work.

Rotations and codebooks are often applied in IVFADC variations, but there are significant costs associated with them. In the most extreme cases, Locally Optimized Product Quantization (LOPQ)~\cite{LOPQ} learns a separate rotation matrix and codebook for each partition. This leads to an extra memory cost of $O(m d (d+n_W))$ and $O(m_{ADC} d (d+n_W))$ more multiplications for each query at search time.
where
$m_{ADC}$ is the number of VQ partitions we search.
When $m$ and $m_{ADC}$ increase, the overhead become quickly noticeable and may become even more expensive than ADC itself. For example, when $d=200$, $n_B=50$, $n_W=16$, performing the rotation once is as expensive as performing 6,400 ADC computations under an optimized implementation. In practice, it is often desirable to avoid per partition rotation or codebooks, but learn global codebooks and rotation.

\section{Methods}

Existing approaches based on the IVFADC framework mainly focus on minimizing the squared loss when quantizing residuals.
Formally, they aim at finding an optimal quantization parameter $\theta^* = \argmin_\theta \frac{1}{|\bX|}\sum_{x \in \bX} \|r_x - \phi(r_x; \theta)\|_2^2$.
As we have discussed in previous sections, our ``signal'', i.e., residual IPs $q \cdot r_x$ exhibit strong non-uniformity locally within a partition.
By directly taking advantage of this skewed distribution, our proposed method achieves higher recall at the same bitrate when compared with others that are agnostic of this phenomenon.

\subsection{Local Orthogonal Decomposition}

Given a unit norm vector or direction $\|v\|_2 = 1$, we define:
\[
H^\parallel_v = vv^T, \, H^\perp_v = I - H^\parallel_v
\]
Hence $H^\parallel_v$ is the projection matrix onto direction $v$ and $H^\perp_v$ is projection matrix onto its complement subspace.
We can thus decompose a residual as: $r_x = H^\parallel_v r_x + H^\perp_v r_x$.

Similar to the original IVFADC framework, we first decompose the IP between a query $q$ and a database vector $x$ into $q \cdot x = q \cdot c + q \cdot r_x$.
With our new insight of non-uniformity of the distribution of our signal, we propose to further decompose the residual IP with respect to a learned direction $v$ as:
\[
q \cdot r_x = q \cdot (H^\parallel_v r_x) + q \cdot (H^\perp_v r_x)
\]

We name  $H^\parallel_v r_x$ the projected component of residual $r_x$ and $H^\perp_v r_x$ the orthogonal component.
Note that the projected component resides in a 1-dimensional subspace spanned by $v$ and can also be very efficiently quantized with existing scalar quantization techniques.

\subsection{Multiscale Quantization of Orthogonal Component}

We define $o^v_x = H^\perp_v r_x$ and $\hat{o^v_x} = o^v_x / \|o^v_x\|_2$ to simplify notation.
Multiscale quantization proposed in \cite{MSQ} learns a separate scale and rotation matrix that are multiplied to the product quantized residual as $\lambda_x R \phi_{PQ}(\hat{o^v_x})$, where $R$ is a learned rotation matrix and $\phi_{PQ}(\cdot)$ is the production quantization learned from the normalized orthogonal components.
Differently from the original MSQ, our scale is chosen to preserve the $\ell_2$ norm of the orthogonal component $o^v_x$, not the whole residual $r_x$:
\[
\| \lambda_x H^\perp_v \phi_{PQ}(\hat{o^v_x}) \|_2 = \| o^v_x \|_2
\]
The rotation $R$ is omitted as it doesn't affect the $\ell_2$ norm.
Another scalar quantization (SQ) is learned on the scales to further reduce the storage cost and speedup ADC.
The final MSQ quantized residual is then:
\[
\phi_{MSQ}(o^v_x) = \phi_{SQ}(\lambda_x) R \phi_{PQ}(\hat{o^v_x})
\]
Where $\phi_{SQ}$ is the non-uniform scalar quantization for partition $P$ learned via a Lloyd algorithm.
The number of codewords in the SQ codebook is fixed at 16 for our experiments, hence its storage cost is negligible.

\subsection{Adjustment to Projected Component}

In general, unlike $o^v_x$, $\phi_{MSQ}(o^v_x)$ is \emph{not} orthogonal to $v$ anymore.
Recall that we want to approximate $q \cdot o^v_x$ in the orthogonal subspace as $q \cdot (H^\perp_v \phi_{MSQ}(o^v_x))$.
Now a subtle performance issue arises.
A critical improvement to ADC introduced since the original OPQ~\cite{OPQ} is to move the rotation multiplication to the query side so that it is done only once globally.
Formally with MSQ, we can perform following: $q \cdot \phi_{MSQ}(r_x) = q \cdot (\phi_{SQ}(\lambda_x) R \phi_{PQ}(\hat{o^v_x})) = \phi_{SQ}(\lambda_x)((R^Tq) \cdot \phi_{PQ}(\hat{o^v_x}))$.

With LOD, the extra projection $H^\perp_v$ in front of $\phi_{MSQ}(o^v_x)$ prevents us from moving $R$ to the $q$ side as the two matrices $H^\perp_v$ and $R$ are \emph{not} commutative in general.
However we have $q \cdot o^v_x \approx q \cdot H^\perp_v \phi_{MSQ}(o^v_x) = (q \cdot \phi_{MSQ}(o^v_x)) - (q \cdot (H^\parallel_v \phi_{MSQ}(o^v_x)))$.
We can perform fast ADC on the term $q \cdot \phi_{MSQ}(o^v_x)$ as proposed in the orginal MSQ~\cite{MSQ} and only multiply matrix $R^T$ to $q$ once.
The extra term $ q \cdot (H^\parallel_v \phi_{MSQ}(o^v_x)) = (q \cdot v)(\phi_{MSQ}(o^v_x) \cdot v)$ can be removed by subtracting it from the projected component before quantization.

\subsection{Uniform Quantization of Projected Component}

Following the procedure above, after projection onto direction $v$, we have the original residual contributing $r_x \cdot v$ and its quantized orthogonal component contributing an extra term $\phi_{MSQ}(o^v_x) \cdot v$.
We thus need to quantize the difference between the two as $z^v_x = (r_x - \phi_{MSQ}(o^v_x)) \cdot v$.
We propose to learn a uniform quantization:
\[
\phi_{UQ}(z^v_x; a_P, b_P) = a_P \textrm{round}((z^v_x - b_P) / a_P) + b_P
\]
Whereby:
\begin{itemize}
    \item $z_{\max}$ and $z_{\min}$ are the maximum and minimum of the finite input set $\{ z^v_x | x \in P \}$. And $l_{UQ}$ is the number of bits for uniform quantization;
    \item $a_P = (z_{\max} - z_{\min}) / (2^{l_{UQ}} - 1)$ scales the input into the range $[-2^{l_{UQ} - 1}, +(2^{l_{UQ} - 1} - 1)]$.
    \item $b_P = (z_{\max} + z_{\min} + a_P) / 2$ centers the input;
    \item $\textrm{round}(\cdot)$ is the function that rounds a floating point number to its nearest integer.
\end{itemize}

$\textrm{round}((z^v_x - b_P) / a_P)$ is the integer code in $l_{UQ}$ bits that we store for each residual.
In practice, we may relax $z_{\max}$ to the $99\%$th quantile of the input to guard against outliers, and similarly $1\%$th quantile for $z_{\min}$.
We clip rounded outputs to within $[-2^{l_{UQ} - 1}, +(2^{l_{UQ} - 1} - 1)]$.

The main advantage of UQ over other scalar quantization techniques is that its codebook size is independent of the number of bits used for its codes.
This is critical as we use $l_{UQ} = 256$ for our experiments.
It also enables fast computation of approximate IP between query and projected component as: $(q \cdot v) \phi_{UQ}(z^v_x) = ((q \cdot v) a_P) \textrm{round}((z^v_x - b_P) / a_P) + (q \cdot v) b_P$.

\begin{comment}
\subsection{Partition Selection}

Given a query $q$ and the set $C$ of centers of all partitions, we propose to pick the partition whose center has the maximum IP with $q$, i.e.,
\[
c^* = \argmax_{c \in C} q \cdot c
\]
Once we choose $c^*$ via this maximum inner product criterion, we then proceed to apply ADC in partition $P^*$ whose center is $c^*$.
\end{comment}

Putting both quantization schemes together, we can approximate the residual IP by replacing each component with its quantized result:
\begin{eqnarray*}
& q \cdot r_x = q \cdot (H^\parallel_v r_x) + q \cdot o^v_x & \\
& \approx (q \cdot v) \phi_{UQ}(z^v_x) + q \cdot \phi_{MSQ}(o^v_x) &
\end{eqnarray*}
And for each term, we can perform efficient ADC.

\begin{comment}
For the first item:
\begin{eqnarray*}
& (q \cdot v) \phi_{UQ}(z^v_x) = & \\
& ((q \cdot v) a_P) \textrm{round}((z^v_x - b_P) / a_P) + (q \cdot v) b_P &
\end{eqnarray*}
And for the second:
\begin{eqnarray*}
& q \cdot \phi_{MSQ}(o^v_x) = q \cdot (\phi_{SQ}(\lambda_x) R \phi_{PQ}(\hat{o^v_x})) & \\
& = \phi_{SQ}(\lambda_x)((R^T q) \cdot \phi_{PQ}(\hat{o^v_x})) &
\end{eqnarray*}
\end{comment}

\subsection{Preserving $\ell_2$ Norms}

We design the LOD+MSQ framework with the objective of preserving $\ell_2$ norms of residuals.
Note that:
\begin{eqnarray*}
& (q \cdot v)\phi_{UQ}(z^v_ x) + q \cdot\phi_{MSQ}(o^v_x) = & \\
& q \cdot \big((\phi_{UQ}(z^v_ x) v + H^\parallel_v \phi_{MSQ}(o^v_x)) + H^\perp_v \phi_{MSQ}(o^v_x)\big) &
\end{eqnarray*}

In the projected subspace, we have:
\begin{eqnarray*}
& (\phi_{UQ}(z^v_ x) v + H^\parallel_v \phi_{MSQ}(o^v_x) \approx & \\
& z^v_x v + H^\parallel_v \phi_{MSQ}(o^v_x) = r_x \cdot v&
\end{eqnarray*}
In the orthogonal subspace, we have 
\begin{eqnarray*}
& \|H^\perp \phi_{MSQ}(o^v_x)\|_2^2 =  \|\phi_{SQ}(\lambda_x) H^\perp_v \phi_{PQ}(\hat{o^v_x}) \|_2 \approx & \\
& \|\lambda_x H^\perp_v \phi_{PQ}(\hat{o^v_x}) \|_2 = \| o^v_x \|_2
\end{eqnarray*}
Hence we preserve the $\ell_2$ norm of $r_x$ up to small scalar quantization errors in $\phi_{UQ}(z^v_ x)$ and $\phi_{SQ}(\lambda_x)$.
Empirically, preserving $\ell_2$ norms improves recall when there is considerable variation in residual norms \cite{MSQ}.

\subsection{Indexing and Search Algorithms}

We list all parameters of the overall indexing and search algorithms besides their inputs in Table~\ref{tbl:indexingparams}.

\begin{table}[ht]
\begin{tabular}{ |p{0.9cm} | p{6.cm}| }
\hline
$m$ & \#partitions in the inverted file \\
$n_B$ & \#codebooks used for PQ encoding \\
$n_W$ & \#codewords used in each PQ codebook \\
$l_{UQ}$ & \#bits for UQ encoding \\
$l_{SQ}$ & \#bits for SQ encoding \\
$m_{ADC}$ & \#partitions to apply ADC to \\
\hline
\end{tabular}
\caption{Parameters for the overall indexing and search algorithms.}
\label{tbl:indexingparams}
\end{table}

\begin{algorithm}[ht]
\begin{small}
\SetKwFunction{FFIndex}{Index}
\SetKwFunction{FFProjectionDirection}{ProjDir}
\SetKwFunction{FFIVF}{IVF}
\SetKwFunction{FFOptimizedPQ}{OPQ}
\SetKwFunction{FFScalarQuantize}{ScalarQuantize}
\SetKwFunction{FFUniformQuantize}{UniformQuantize}
\SetKwInOut{Input}{input}
\SetKwInOut{Output}{output}
\DontPrintSemicolon

\FFIndex{$\bX$} \Begin {
  \Input{Database $\bX$ and function \FFProjectionDirection}
  \Output{Partitions $\{P^i\}_{[m]}$ with centers $\{c^i\}_{[m]}$, projection directions $\{v^i\}_{[m]}$, uniform quantization $\{\phi_{UQ}^i(\cdot)\}_{[m]}$ and multiscale quantization $\{\phi_{MSQ}^i(\cdot)\}_{[m]}$}
  \BlankLine
  $\{P^i\}_{[m]}, \, \{c^i\}_{[m]} \ \leftarrow  \FFIVF(\bX, m)$ \;
  \For{$i \leftarrow 1$ \KwTo $m$} {
    $v^i \leftarrow \FFProjectionDirection(P^i, c^i)$ \;
    Compute $\{ o^v_x \leftarrow H_{v^i}^\perp r_x \}_{x \in P^i}$ \;
    Compute $\{ \hat{o^v_x} \leftarrow o^v_x / \| o^v_x \|_2 \}_{x \in P^i}$ \;
  }
  $R, \, \phi_{PQ}(\cdot) \leftarrow \FFOptimizedPQ(\{ \hat{o^v_x} \}_{x \in \bX}, n_B, n_W )$ \;
  
  \For{$i \leftarrow 1$ \KwTo $m$} {
    Compute $\{ \lambda_x \leftarrow \| o^v_x \|_2 / \| H_{v^i}^\perp \phi_{PQ}(\hat{o^v_x} )\|_2 \}_{x \in P^i}$ \;
    $\phi_{SQ}^i(\cdot) \leftarrow \FFScalarQuantize(\{ \lambda_x \}_{x \in P^i}, l_{SQ})$ \;
    Compute $\{\phi_{MSQ}^i(o^v_x) \leftarrow \phi_{SQ}^i(\lambda_x)R\phi_{PQ}(\hat{o^v_x})\}_{x \in P_i}$ \;
    Compute $\{ z_x \leftarrow (r_x - \phi_{MSQ}^i(o^v_x)) \cdot v^i \}_{x \in P^i}$ \;
    $\phi_{UQ}^i(\cdot) \leftarrow \FFUniformQuantize(\{ z_x \}_{x \in P^i}, l_{UQ})$ \;
  }
  \Return $\{P^i\}_{[m]}, \, \{c^i\}_{[m]}, \, \{v^i\}_{[m]}, \{\phi_{UQ}^i(\cdot)\}_{[m]}, \, \{\phi_{MSQ}^i(\cdot)\}_{[m]}$
}
  \caption{Index database $\bX$ with local orthogonal decomposition and multiscale quantization. The projection direction is parameterized with the function \texttt{ProjDir}.} 
\label{algo:index_database}
\vspace{-0.02in}
\end{small}
\end{algorithm}

\begin{algorithm}[ht]
\begin{small}
\SetKwFunction{FFIndex}{Index}
\SetKwFunction{FFSearch}{Search}
\SetKwFunction{FFTop}{Top}
\SetKwFunction{FFADC}{ADC}
\SetKwFunction{FFOptimizedPQ}{OPQ}
\SetKwFunction{FFScalarQuantize}{ScalarQunatize}
\SetKwFunction{FFUniformQuantize}{UniformQuanitze}
\SetKwInOut{Input}{input}
\SetKwInOut{Output}{output}
\DontPrintSemicolon

\FFSearch{$q, \, k$} \Begin {
  \Input{query $q$, number $k$ and outputs of $\FFIndex(\bX)$}
  \Output{Approximate top $k$ maximum inner products}
  \BlankLine
  Compute $\{ p_i \leftarrow q \cdot c^i \}_{[m]}$ \;
  $C^* \leftarrow \FFTop(\{ (p_i, i) \}_{[m]}, \, m_{ADC})$ \;
  $q_r \leftarrow R^T q$ \;
  \For{$(\textrm{p}_i, i ) \in C^*$} {
    Compute $\{ \textrm{rip}^\perp_x \leftarrow \FFADC(q_r, \phi_{PQ}^i(\hat{o^v_x}) ) \}_{x \in P_i}$ \;
    Compute $\{ \textrm{rip}^\perp_x \leftarrow \phi_{SQ}^i(\lambda_x) \times \textrm{rip}^\perp_x \}_{x \in P_i}$ \;
    Compute $\{ \textrm{rip}^\parallel_x \leftarrow (q \cdot v_i) \times \phi_{UQ}^i(z_x) \}_{x \in P_i}$ \;
    Compute $\{ \textrm{rip}_x \leftarrow \textrm{rip}^\perp_x + \textrm{rip}^\parallel_x \}_{x \in P_i}$ \;
    \tcp{$i_x$: index of $x$}
    $\textrm{top}_i \leftarrow \FFTop(\{ (\textrm{rip}_x, i_x) \}_{x \in P_i}, k)$ \;
    $\textrm{top}_i \leftarrow \{ (\textrm{rip}_x + p_i, i_x) \}_{(\textrm{rip}_x, i_x) \in \textrm{top}_i}$ \;
  }
  \Return $\FFTop(\cup_{(p_i, i) \in C^*} \textrm{top}_i, k)$
}
\caption{Search top $k$ inner products in an indexed database with query $q$.} 
\label{algo:search_database}
\end{small}
\end{algorithm}

We want to highlight that in memory bandwidth limited large scale MIPS, the search time is well approximated by the number of bits read: $O(N\frac{m_{ADC}}{m}(l_{UQ} + n_B \lceil \log_2 n_W \rceil))$.
In our experiments, we fix $m_{ADC}/m = 1/10$.
The bitrate of the original dataset is 32 bits per dimension and we use either $1/2$ or 1 bit per dimension in our quantization schemes.
Hence we achieve over 2-orders of magnitudes of speedup.

\section{Analysis}

We leave the projection direction function as an input to our indexing algorithm in the previous section.
In this section, we formally investigate the optimal projection direction given partition $P$ and its center $c$ conditional on the fact that $c = \argmax_{c_i \in C} q \cdot c_i$.

We start by analyzing the error introduced by our quantization scheme to the approximate residual IP.
Let $e_{UQ}(z^v_x) = (q \cdot v)(z^v_ x - \phi_{UQ}(z^v_x))$ and $e_{MSQ}(o^v_x) = (H^\perp_v q) \cdot (o^v_x - \phi_{MSQ}(o^v_x))$.
Consider the quantization error on the residual IP within partition $P$ as:
\begin{eqnarray*}
& \frac{1}{|P|} \sum_{x \in P} (q \cdot r_x - (q \cdot v)\phi_{UQ}(z^v_x) - q \cdot \phi_{MSQ}(o^v_x))^2 = & \\
& \frac{1}{|P|} \sum_{x \in P}  (e_{UQ}(z^v_x) + e_{MSQ}(o^v_x))^2 = & \\
& \frac{1}{|P|} \sum_{x \in P}  (e_{UQ}(z^v_x)^2 + 2 e_{UQ}(z^v_x)e_{MSQ}(o^v_x) + e_{MSQ}(o^v_x)^2) &
\end{eqnarray*}
First, UQ achieves an error bound of $O(2^{-l_{UQ}})$ in its 1-dimensional subspace, which is much lower than the error bound that MSQ can achieve in the orthogonal $(d-1)$-dimensional subspace.
UQ and MSQ are two completely separate quantization steps, and the cross product of their quantization errors are expected to be small.
Therefore we shall focus on minimizing the last quantization error term averaging over $q$ and $r_x$:
\begin{eqnarray*}
& \frac{1}{|P|} \E_q \sum_{x \in P}((H^\perp_v q) \cdot e_{MSQ}(o^v_x))^2 = & \\
& \E_q (H^\perp_v q)^T (\frac{1}{|P|} \sum_{x\in P} e_{MSQ}(o^v_x) e_{MSQ}(o^v_x)^T) H^\perp_v q &
\end{eqnarray*}
If we define  $\Sigma_v = \frac{1}{|P|} \sum_{x \in P } (e_{MSQ}(o^v_x)e_{MSQ}(o^v_x)^T)$ and $o^v_q = H^\perp_v q$, we can then rewrite the optimization as: $\min_v \E_q (o^v_q)^T \Sigma_v  o^v_q$.
Notice that the matrix $\Sigma_v$ in the middle is also dependent on the direction $v$, which makes this optimization problem very challenging.

However the learned rotation $R$ in MSQ serves two purposes: 1) it reduces correlation between dimensions and 2) it evens variance allocation across PQ subspaces \cite{OPQ}.
Hence it is reasonable to expect the errors to be close to isotropic across dimensions assuming the subspace spanned by orthogonal components does not degenerate into a low dimensional subspace.
This is to assume:
\begin{assumption}
The empirical covariance matrix of orthogonal component errors $\{ e_{MSQ}(o^v_x) \}_{x \in P}$ is isotropic.
\end{assumption}

This assumption allows us to approximate $\Sigma_v \approx \lambda I$ with some constant $\lambda$.
Now we arrive at $\min_v \E_q \|H^\perp_v q \|_2^2 = \min_v \E_q q^T(I - vv^T) q = \min_v \E_q q^Tq - \E_q (q \cdot v)^2$.
Let's introduce a simplfication of the conditional expectation as $\E_q(\cdot |c) = \E_q(\cdot | c = \argmax_{c_i \in C} q \cdot c_i)$.
We need to solve the maximization problem of: $\max_v \E_q (q \cdot v)^2 = \max_v v^T \E_q(qq^T|c) v$.
The matrix in the middle is the conditional covariance matrix of all queries that have maximum IPs with center $c$.
If we can estimate this matrix $\E_q (qq^T|c)$ accurately, we can simply take its first principal direction as our optimal direction $v$.

In real applications, for any partition center, we can only sample a very limited number of queries $q$ such that $c = \argmax_{c_i \in C} q \cdot c_i$.
This approach thus can't scale to large $m$ in the range of $10^3-10^6$.
This makes the estimation of $\E_q(qq^T|c)$ inherently of high variance.
To overcome this noisy estimation issue, we provide both theoretical and empirical support of approximating the optimal direction with the partition center direction $u = c/\|c\|_2$.

\subsection{Alignment of Query and Partition Center}

We first estimate the magnitude of the projected query component along the partition center direction.
In the original setup, we have a set of fixed centers and a random query.
To facilitate our analysis, we can fix the query and instead \emph{rotate} centers with respect to the query.
We start by studying the case where both centers and query are normalized and later lift the constraint.
We consider the scenario where centers after rotation follow a uniform distribution over the unit sphere $\mathbf{S}_{d-1}$.
This provides a more conservative bound than that of real datasets, because real queries tend to be tightly clustered around the ``topics'' in the database due to formulation of the training objectives and regularizers \cite{SpreadoutZhang}.

\begin{theorem}
Given a normalized query $q \in \bR^d$ and $m$ random centers $C = \{ c_i \}_{[m]}$ uniformly sampled from the unit sphere $\mathbf{S}_{d-1}$,
with probability at least $1-\delta$, the maximum cosine similarity between the query and $c^* = \argmax_{c \in C} q \cdot c$ is at least $L_1(m, \delta)$:
\begin{eqnarray*}
& Pr\{ \cos(q, c^*) \ge L_1(m, \delta) \} \ge 1 - \delta & \\
& L_1(m, \delta) = \sqrt{1 - \big(\frac{\eta_1 \sqrt{d} \log 1/\delta}{m} \big)^{\frac{2}{d + 1}}} &
\end{eqnarray*}
In practical settings, we have $\log(m/(\eta_1 \sqrt{d} \log 1/\delta)) \ll (d+1)/2$. Let $\alpha = 2(1-e^{-1}) > 1$, we can weaken it to a more intuitive form:
\[
L_1(m, \delta) \ge \sqrt{\alpha \max \big( \frac{\log (m / \sqrt{d})- \log(\eta_1 \log 1/\delta)}{d +1}, 0 \big)}
\]
\label{thm:random_centers}
\end{theorem}

\begin{lemma}
If we uniformly sample 2 vectors $x$ and $y$ from the unit sphere $\mathbf{S}_{d-1}$, we have $\E_{x, y} (x \cdot y)^2 = 1/d$
\end{lemma}

A few comments on these 2 results:
\begin{itemize}
    \item From Theorem~\ref{thm:random_centers}, we can see that the dependency of the maximum residual IP on the confidence parameter $\delta$ is rather weak at $\log\log(1/ \delta)$.
    \item If we choose $\delta = 1/2$, we can thus show that for at least half of the queries, the largest IP $q \cdot c^*$ is at least $O(\sqrt{\log (m / \sqrt{d})})$ larger than the cosine similarity between two randomly sampled vectors.
\end{itemize}

Next, we allow centers to have varying norms:

\begin{theorem}
Suppose the directions of $m$ centers $C = \{ c_i \}_{[m]}$ are uniformly sampled from the unit sphere $S_{d-1}$, and their sorted norms are $h_1 \ge h_2 \ge \cdots \ge h_m$.
With probability at least $1-\delta$, the maximum cosine similarity between the query and $c^* = \argmax_{c \in C} q \cdot c$ is at least $L_2(m, \delta, \{h_i\}_{[m]})$:
\[
L_2(m, \delta, \{h_i\}_{[m]}) = \max_{i \in [m]} \frac{h_i}{h_1} L_1(i, \delta)
\]
\label{thm:random_centers_varying_norm}
\end{theorem}
\vspace{-0.2in}

Intuitively, as $i$ increases, the first factor $\frac{h_i}{h_1}$ decreases, but the second one $L_1(i, \delta)$ increases, thus the maximum is achieved somewhere in the middle.
Specifically, we can see that $L_2(m, \delta, \{h_i\}_{[m]}) \ge \frac{h_{\lceil m/2 \rceil}}{h_1} L_1(\lceil m/2 \rceil, \delta)$.
This bound is robust to any small outlier near $h_m$, but it can be influenced by the largest norm $h_1$.

However, we remark that when the largest center norm $h_1$ is significantly larger than the median $h_{\lceil m/2 \rceil}$, the MIPS problem itself becomes much easier.
As the relative magnitude of $h_1$ increases, its partition becomes more likely to contain the maximum IP than the rest.
And furthermore, the gap between the maximum IP in $h_1$'s partition and the maximum IP from other partitions becomes wider.
Both the concentration of the maximum IP in one partition and the large gap contribute to better recall.
Hence LOD helps adversarial instances \emph{more} than easy instances, which explains the consistent recall improvement in our experiments.
Exact quantification of this behavior is one of our future research directions.

We conclude this section with the observation that real queries tend to be \emph{more} clustered along partition centers than what is suggested by Theorem~\ref{thm:random_centers}, i.e., the observed $\cos^2(q, c^*)$ is much higher than $O(\log (m/\sqrt{d}) / d)$.
We hypothesize that this is due to the training process that aligns query vectors with the natural topical structure in the database vector space.

\subsection{Asymptotically Optimal Projection Direction}

Let $o^u_q = H^\perp_u q$ be the orthogonal query component in the complement subspace of the partition center.
Under the same assumption as Theorem~\ref{thm:random_centers_varying_norm}, we are ready to state our main result:
\begin{theorem}
Let $\gamma$ be the ratio between the the largest and smallest non-zero eigenvalues of the matrix $\E_q (o^u_q(o^u_q)^T |c)$.
The optimal direction is equal to the partition center direction with probability at least $1-\delta$ if:
\[
\gamma < (d-2)L_2^2(m, \delta, \{h_i\}_{[m]})
\]
With some positive constant $\eta_2$, we can rewrite the above into a more intuitive form:
\[
\gamma < \eta_2 (\log (m / \sqrt{d}) - \log(\eta_1 \log 1/\delta))
\]
\label{thm:optimal_direction}
\end{theorem}
\vspace{-0.27in}
This theorem states that when the number of partitions $m$ increases above a threshold dependent on the ratio $\gamma$ and $\delta$, the optimal direction is equal to the partition center direction with probability at least $1-\delta$.
Hence asymptotically, the optimal direction approaches the partition center direction for our LOD+MSQ framework as $m \to \infty$ and $m \gg d$.

\subsection{Approximation with Benefits}

Approximating the optimal direction with the partition center direction also brings practical benefits:
\begin{itemize}
    \item \textbf{No extra storage cost}, as we don't have to store a separate vector per partition.
    \item \textbf{Free projection at search time}, as we have computed all IPs between the query and centers for partition selection. We just need to perform an $O(1)$ operation to divide the IP by the center norm to get the projected component $q \cdot (c /\|c\|_2$). 
\end{itemize}

\section{Experiments}
\label{sec:experiments}

\subsection{Datasets}

We apply our method along with other state-of-the-art MIPS techniques on public datasets Netflix \cite{BennettNetflixPrize} and Glove~\cite{PenningtonGlove}.
The Netflix dataset is generated with a regularized matrix factorization model similar to \cite{GreedyMIPSYu}.
The Glove dataset is downloaded from \texttt{https://nlp.stanford.edu/projects/glove/}.
For word embeddings, the cosine similarity between these embeddings reflects their semantic relatedness.
Hence we $\ell_2$-normalize the Glove dataset, and then cosine similarity is equal to inner product.

We list details of these datasets in Table~\ref{tbl:datasets}.

\begin{table}[ht]
\centering
\begin{tabular}{| l | l | l |}
\hline
Dataset & \#Vectors & \#Dims \\
\hline
Netflix & 17,770 & 200 \\
Glove & 1,183,514 & 200 \\
\hline
$m$ & $m_{ADC}$ & $m_{ADC} / m$ \\
\hline
20 & 2 & 10\% \\
1000 & 100 & 10\% \\
\hline
\end{tabular}
\caption{Datasets used for MIPS experiments.}
\label{tbl:datasets}
\end{table}

\subsection{Recalls}

We apply following algorithms to both of our datasets:

\begin{itemize}

\item \textbf{MIPS-PQ}: implements the PQ \cite{PQ} quantization scheme proposed in the original IVFADC framework.

\item \textbf{MIPS-OPQ}: implements the OPQ \cite{OPQ} quantization scheme that learns a global rotation matrix.

\item \textbf{L2-OPQ}: implements the OPQ quantization scheme and also the MIPS to $\ell_2$-NNS conversion proposed in \cite{MIPS-to-L2}. We do not transform the Glove dataset since $\ell_2$-NNS retrieves the same set of database vectors as MIPS after normalization.

\item \textbf{MIPS-LOD-MSQ}: implements our proposed method with both LOD and MSQ. The projection direction is set to the partition center as an effective approximation to the optimal direction.

\end{itemize}

We set parameters to following values for all our recall experiments:

\begin{itemize}

\item \textbf{IVF}: we keep average partition size at around 1,000 and we always search 10\% of the partitions with ADC. This is in-line with other practices reported in benchmarks and industrial applications \cite{Deep10m, FAISS}.

\item \textbf{Product Quantization}: we use either $n_B \in \{25, 50\}$ codebooks, each of which contains $n_W = 16$ codewords for PQ and OPQ. For LOD+MSQ, we set $n_B \in \{23, 48\}$ when $l_{UQ}=8$ and $n_B \in \{24, 49\}$ when $l_{UQ}=4$ to keep the number of bits spent on each database vector the same. The number of codewords is fixed at 16 for efficient SIMD based implementation of in-register table look-up~\cite{MSQ, BlalockBOLT}.

\item \textbf{UQ}: we use $l_{UQ} = 8$ bits for uniform quantization for Netflix and $l_{UQ} = 4$ bits for Glove, which results in 256 and 16 levels in the codebook respectively.

\item \textbf{MSQ}: we use $l_{SQ} = 4$ bits and accordingly 16 levels for scalar quantization of scales in MSQ for all experiments. We apply the same technique in \cite{MSQ} to avoid explicitly storing the codes and hence it incurs no extra cost in storage.

\end{itemize}

The combination of LOD+MSQ consistently outperforms other existing techniques under the same bitrate.
Its relative improvement is higher on Netflix because the residual norms of the Netflix dataset exhibit larger variance than those of the Glove dataset.

\begin{figure}[ht]
\begin{subfigure}[h]{0.48\linewidth}
\includegraphics[width=\linewidth]{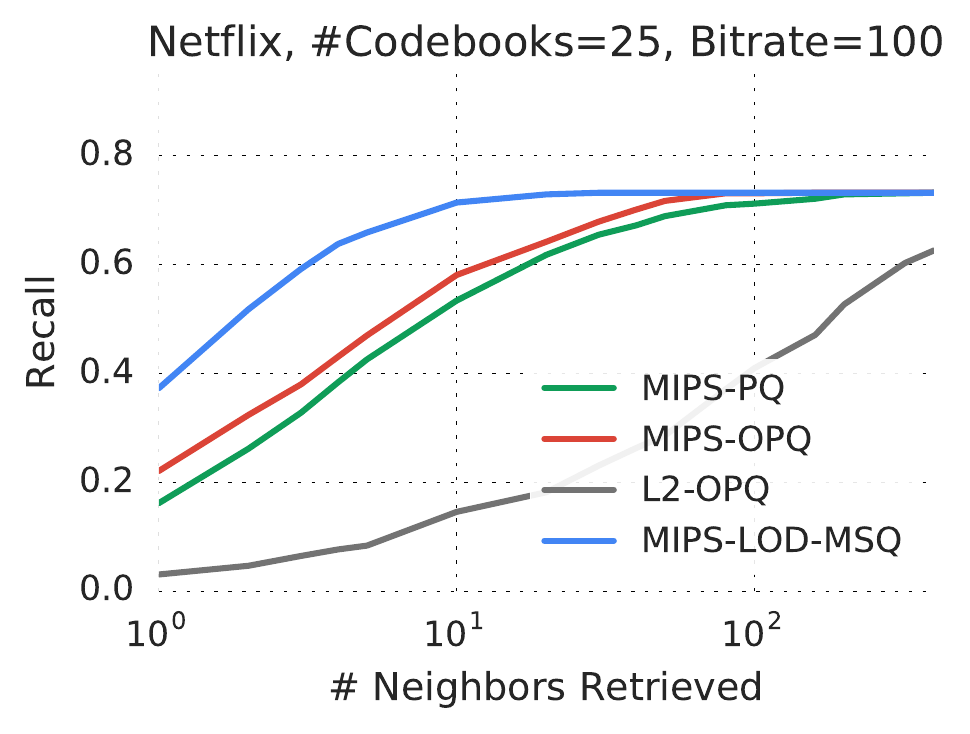}
\caption{}
\label{fig:netflix_100}
\end{subfigure}
\hfill
\begin{subfigure}[h]{0.48\linewidth}
\includegraphics[width=\linewidth]{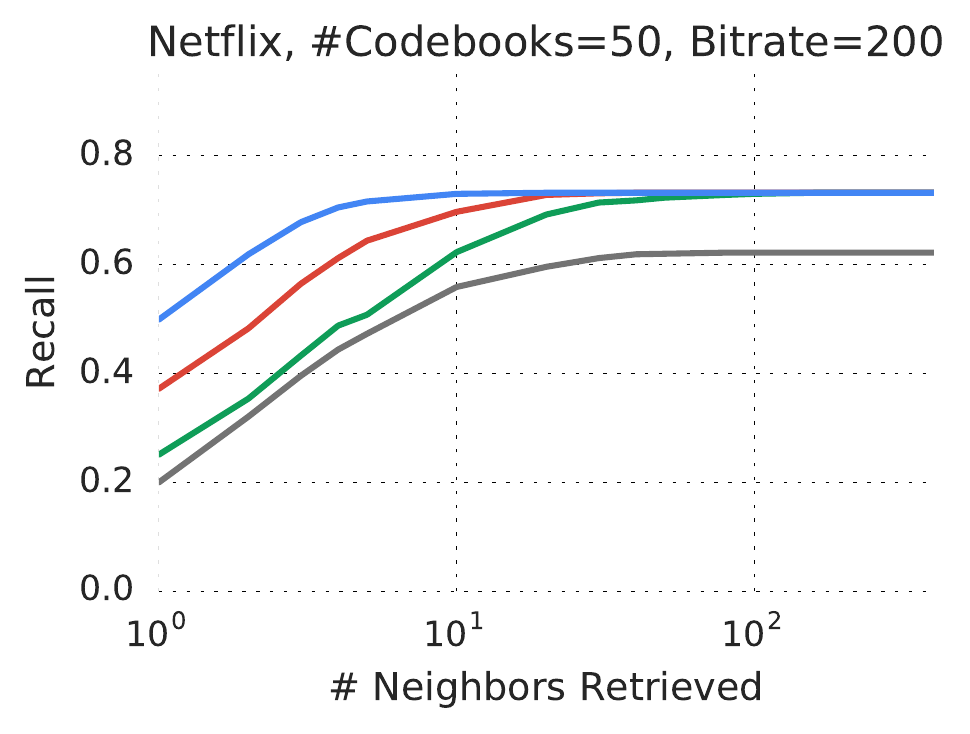}
\caption{}
\label{fig:netflix_200}
\end{subfigure}%
\caption{Experiments on the Netflix dataset: (a) recall vs $k$ for 100-bit encoding of database vectors and (b) recall vs $k$ for 200-bit encoding.}
\end{figure}

\begin{figure}[ht]
\begin{subfigure}[h]{0.48\linewidth}
\includegraphics[width=\linewidth]{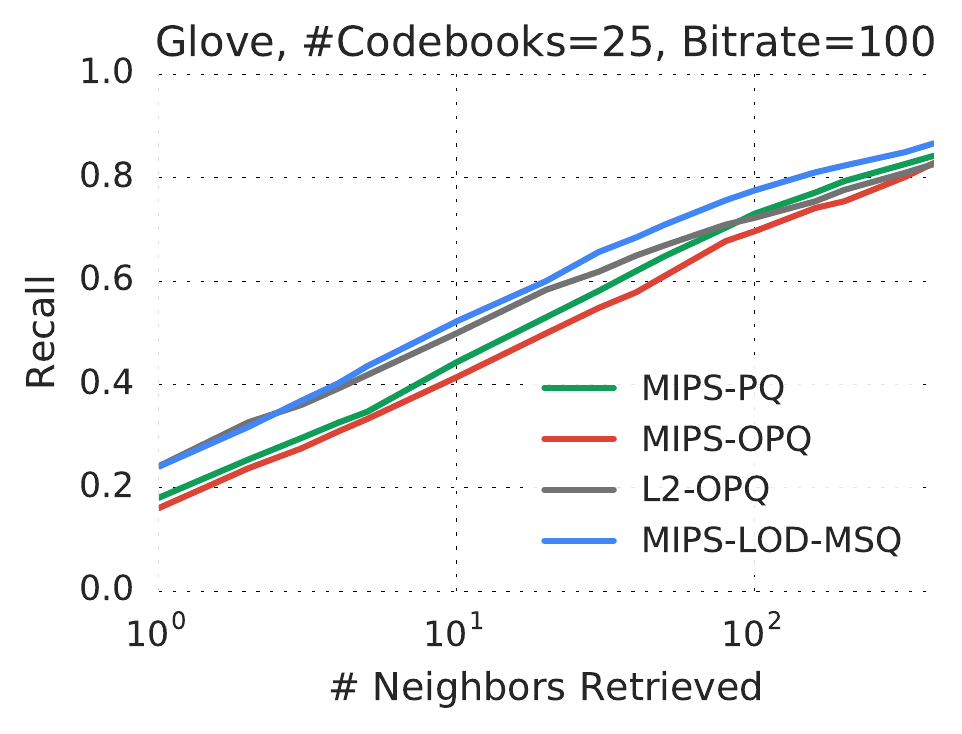}
\caption{}
\label{fig:glove_100}
\end{subfigure}
\hfill
\begin{subfigure}[h]{0.48\linewidth}
\includegraphics[width=\linewidth]{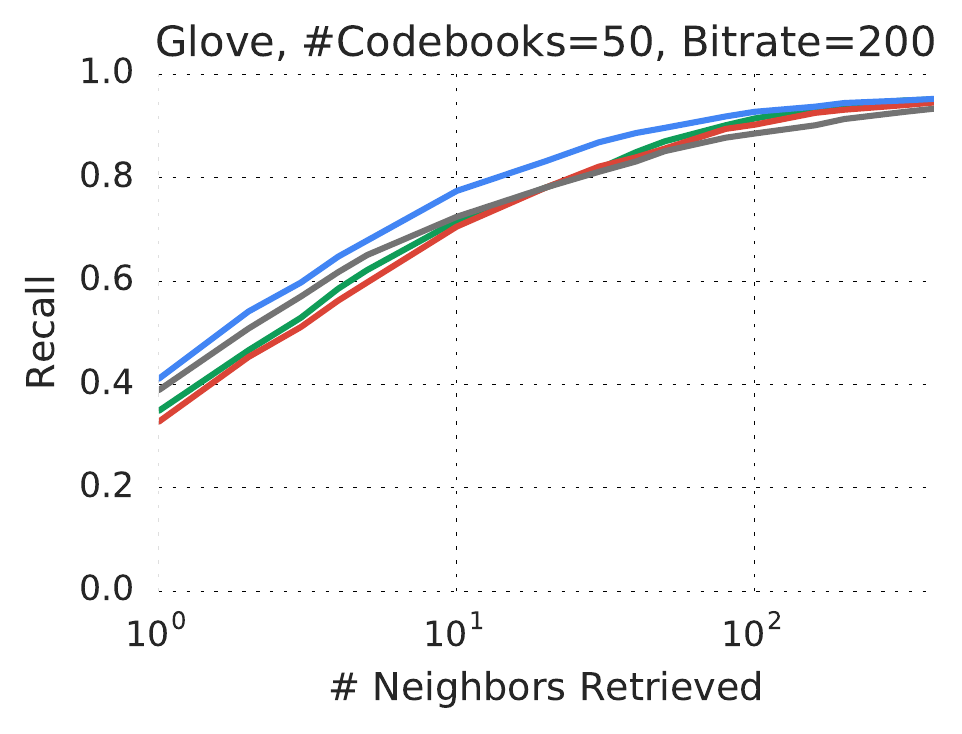}
\caption{}
\label{fig:glove_200}
\end{subfigure}%
\caption{Experiments on the Glove dataset: a) recall vs $k$ for 100-bit encoding of database vector and (b) recall vs $k$ for 200-bit encoding.}
\end{figure}

\begin{figure}[ht]
\begin{subfigure}[h]{0.48\linewidth}
 \includegraphics[width=\linewidth]{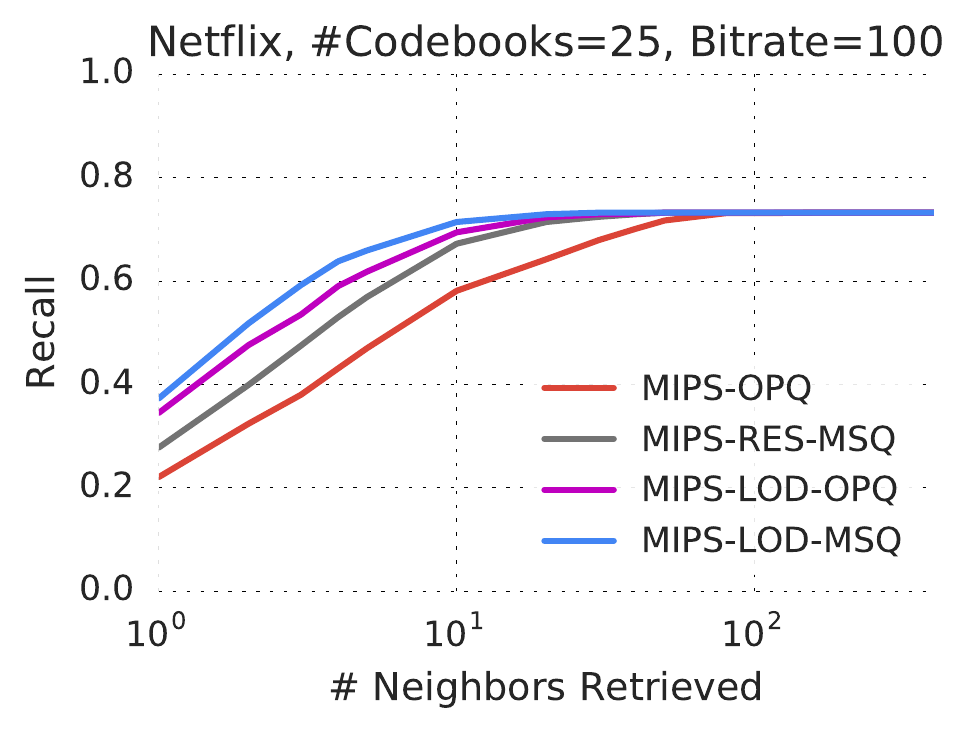}
\caption{}
\label{fig:netflix_ablation}
\end{subfigure}
\hfill
\begin{subfigure}[h]{0.48\linewidth}
\includegraphics[width=\linewidth]{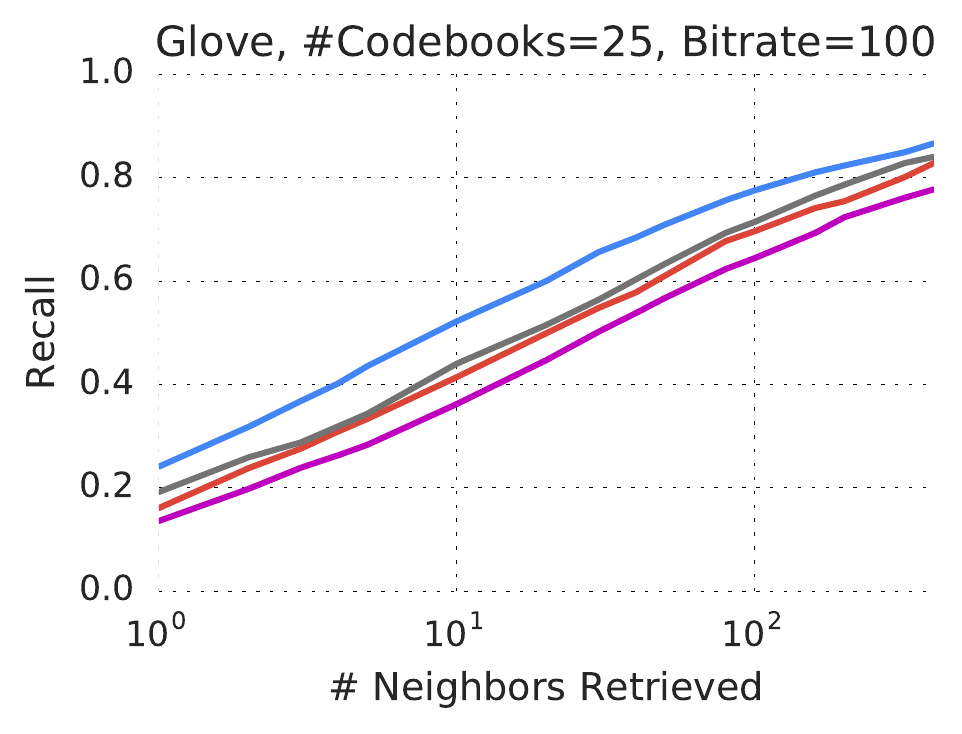}
\caption{}
\label{fig:glove_ablation}
\end{subfigure}%
\caption{Ablation study of both LOD and MSQ on Netflix and Glove. All plots are generated with 100 bit per database vector.}
\end{figure}

\subsection{Ablation}

To systematically investigate the contribution of LOD and MSQ in isolation, we perform ablation study with both datasets.

\begin{itemize}

\item \textbf{MIPS-OPQ, MIPS-LOD-MSQ}: are repeated from the experiments reported from the previous section.

\item \textbf{MIPS-MSQ}: implements the MSQ quantization scheme directly on the residuals $r_x$ without LOD.

\item \textbf{MIPS-LOD-OPQ}: first applies LOD and then implements the OPQ quantization scheme on the orthogonal component $o^v_x$.

\end{itemize}

The combination of LOD+MSQ consistently outperforms either one in isolation. Interestingly, LOD performs much better than MSQ alone on Netflix and worse on Glove.
This is due to the fact that in the normalized Glove dataset, orthogonal components of residuals have larger norms than projected components.
With LOD only, OPQ is applied to the orthogonal components
and it fails to preserve $\ell_2$ norms at a low bitrate.
And the decrease in recall is fairly discernable from the Figure~\ref{fig:glove_ablation}.

\section{Conclusion}

In this work, we propose a novel quantization scheme that decomposes a residual into two orthogonal components with respect to a learned projection direction.
We then apply UQ to the projected component and MSQ to the orthogonal component respectively.
We provide theoretical and empirical support of approximating the optimal projection direction with the partition center direction, which does not require estimating the noisy conditional covariance matrix.
The combination of local orthogonal decomposition and MSQ consistently outperforms other quantization techniques on widely tested public datasets.

\bibliographystyle{abbrv}
\bibliography{paper} 

\begin{thebibliography}{10}

\bibitem{SIMDLUT}
F.~Andr{\'e}, A.-M. Kermarrec, and N.~Le~Scouarnec.
\newblock Cache locality is not enough: high-performance nearest neighbor
  search with product quantization fast scan.
\newblock {\em Proceedings of the VLDB Endowment}, 9(4):288--299, 2015.

\bibitem{KMeansForMIPS}
A.~Auvolat, S.~Chandar, P.~Vincent, H.~Larochelle, and Y.~Bengio.
\newblock Clustering is efficient for approximate maximum inner product search.
\newblock {\em CoRR}, abs/1507.05910, 2015.

\bibitem{AQ}
A.~Babenko and V.~Lempitsky.
\newblock Additive quantization for extreme vector compression.
\newblock In {\em Computer Vision and Pattern Recognition (CVPR), 2014 IEEE
  Conference on}, pages 931--938. IEEE, 2014.

\bibitem{TreeAQ}
A.~Babenko and V.~Lempitsky.
\newblock Tree quantization for large-scale similarity search and
  classification.
\newblock In {\em Proceedings of the IEEE Conference on Computer Vision and
  Pattern Recognition}, pages 4240--4248, 2015.

\bibitem{Deep10m}
A.~Babenko and V.~Lempitsky.
\newblock Efficient indexing of billion-scale datasets of deep descriptors.
\newblock In {\em 2016 IEEE Conference on Computer Vision and Pattern
  Recognition (CVPR)}, pages 2055--2063, June 2016.

\bibitem{MIPS-to-L2}
Y.~Bachrach, Y.~Finkelstein, R.~Gilad-Bachrach, L.~Katzir, N.~Koenigstein,
  N.~Nice, and U.~Paquet.
\newblock Speeding up the xbox recommender system using a euclidean
  transformation for inner-product spaces.
\newblock In {\em Proceedings of the 8th ACM Conference on Recommender
  Systems}, pages 257--264, 2014.

\bibitem{BennettNetflixPrize}
J.~Bennett, S.~Lanning, and N.~Netflix.
\newblock The netflix prize.
\newblock In {\em In KDD Cup and Workshop in conjunction with KDD}, 2007.

\bibitem{BlalockBOLT}
D.~W. Blalock and J.~V. Guttag.
\newblock Bolt: Accelerated data mining with fast vector compression.
\newblock In {\em Proceedings of the 23rd {ACM} {SIGKDD} International
  Conference on Knowledge Discovery and Data Mining}, pages 727--735, 2017.

\bibitem{MIPSRecSys}
P.~Cremonesi, Y.~Koren, and R.~Turrin.
\newblock Performance of recommender algorithms on top-n recommendation tasks.
\newblock In {\em Proceedings of the Fourth ACM Conference on Recommender
  Systems}, pages 39--46, 2010.

\bibitem{MIPSForEC}
T.~Dean, M.~Ruzon, M.~Segal, J.~Shlens, S.~Vijayanarasimhan, and J.~Yagnik.
\newblock Fast, accurate detection of 100,000 object classes on a single
  machine: Technical supplement.
\newblock In {\em Proceedings of IEEE Conference on Computer Vision and Pattern
  Recognition}, 2013.

\bibitem{OPQ}
T.~Ge, K.~He, Q.~Ke, and J.~Sun.
\newblock Optimized product quantization.
\newblock {\em IEEE Transactions on Pattern Analysis and Machine Intelligence},
  36(4):744--755, April 2014.

\bibitem{PQ}
H.~Jegou, M.~Douze, and C.~Schmid.
\newblock Product quantization for nearest neighbor search.
\newblock {\em IEEE transactions on pattern analysis and machine intelligence},
  33(1):117--128, 2011.

\bibitem{FAISS}
J.~Johnson, M.~Douze, and H.~J{\'e}gou.
\newblock Billion-scale similarity search with gpus.
\newblock {\em arXiv preprint arXiv:1702.08734}, 2017.

\bibitem{LOPQ}
Y.~Kalantidis and Y.~Avrithis.
\newblock Locally optimized product quantization for approximate nearest
  neighbor search.
\newblock In {\em Computer Vision and Pattern Recognition (CVPR), 2014 IEEE
  Conference on}, pages 2329--2336. IEEE, 2014.

\bibitem{AQRevisited}
J.~Martinez, J.~Clement, H.~H. Hoos, and J.~J. Little.
\newblock Revisiting additive quantization.
\newblock In {\em European Conference on Computer Vision}, pages 137--153.
  Springer, 2016.

\bibitem{SQ}
J.~Martinez, H.~H. Hoos, and J.~J. Little.
\newblock Stacked quantizers for compositional vector compression.
\newblock {\em CoRR}, abs/1411.2173, 2014.

\bibitem{MIPSSampledSoftmax}
S.~Mussmann and S.~Ermon.
\newblock Learning and inference via maximum inner product search.
\newblock In {\em Proceedings of The 33rd International Conference on Machine
  Learning}, volume~48, pages 2587--2596, 2016.

\bibitem{CKmeans}
M.~Norouzi and D.~J. Fleet.
\newblock Cartesian k-means.
\newblock In {\em Proceedings of the IEEE Conference on Computer Vision and
  Pattern Recognition}, pages 3017--3024, 2013.

\bibitem{PenningtonGlove}
J.~Pennington, R.~Socher, and C.~D. Manning.
\newblock Glove: Global vectors for word representation.
\newblock In {\em Empirical Methods in Natural Language Processing (EMNLP)},
  pages 1532--1543, 2014.

\bibitem{NeuralEpisodicControl}
A.~Pritzel, B.~Uria, S.~Srinivasan, A.~P. Badia, O.~Vinyals, D.~Hassabis,
  D.~Wierstra, and C.~Blundell.
\newblock Neural episodic control.
\newblock In {\em Proceedings of the 34th International Conference on Machine
  Learning}, volume~70, pages 2827--2836, 2017.

\bibitem{ALSH}
A.~Shrivastava and P.~Li.
\newblock Asymmetric lsh (alsh) for sublinear time maximum inner product search
  (mips).
\newblock In {\em Advances in Neural Information Processing Systems}, pages
  2321--2329, 2014.

\bibitem{MSQ}
X.~Wu, R.~Guo, A.~T. Suresh, S.~Kumar, D.~N. Holtmann-Rice, D.~Simcha, and
  F.~Yu.
\newblock Multiscale quantization for fast similarity search.
\newblock In {\em Advances in Neural Information Processing Systems 30}, pages
  5745--5755. 2017.

\bibitem{LossDecompHsu}
I.~E.-H. Yen, S.~Kale, F.~Yu, D.~Holtmann-Rice, S.~Kumar, and P.~Ravikumar.
\newblock Loss decomposition for fast learning in large output spaces.
\newblock In {\em Proceedings of the 35th International Conference on Machine
  Learning}, volume~80, pages 5640--5649, 2018.

\bibitem{GreedyMIPSYu}
H.-F. Yu, C.-J. Hsieh, Q.~Lei, and I.~S. Dhillon.
\newblock A greedy approach for budgeted maximum inner product search.
\newblock In {\em Advances in Neural Information Processing Systems 30}, pages
  5453--5462. 2017.

\bibitem{CQ}
T.~Zhang, C.~Du, and J.~Wang.
\newblock Composite quantization for approximate nearest neighbor search.
\newblock In {\em ICML}, number~2, pages 838--846, 2014.

\bibitem{SCQ}
T.~Zhang, G.-J. Qi, J.~Tang, and J.~Wang.
\newblock Sparse composite quantization.
\newblock In {\em Proceedings of the IEEE Conference on Computer Vision and
  Pattern Recognition}, pages 4548--4556, 2015.

\bibitem{SpreadoutZhang}
X.~Zhang, F.~X. Yu, S.~Kumar, and S.~Chang.
\newblock Learning spread-out local feature descriptors.
\newblock In {\em {IEEE} International Conference on Computer Vision, {ICCV}
  2017, Venice, Italy, October 22-29, 2017}, pages 4605--4613, 2017.

\end{thebibliography}

\newpage
\clearpage

\section{Appendix}

\subsection{Proof of Theorem \ref{thm:random_centers}}

Without loss the generality, we can assume the query is fixed at $q=[1, 0, \cdots, 0]$.
Thus the inner product between the query and a center becomes the value of the first dimension of the center, whose distribution is $F(y) = \frac{1}{Z_d} \int_{-1}^{y}(1-x^2)^{(d-1)/2} dx$,
where $Z_d$ is the normalization constant.
Its value is given by $V_d / V_{d-1}$, where $V_d$ is the volume of the $d$-dimensional unit hyperball: $\pi^{d/2} / \Gamma(d/2 + 1)$. 

Ideally, we want to find the maximum $h$ that still satisfies $(F(h))^m \le \delta$.
For any $\delta > 2^{-m}$, it is clear that $h > 0$.
And we have:
\[
\big( 1 - \frac{1}{Z_d} \int_{h}^{1}(1-x^2)^{(d-1)/2} dx \big)^m \le \delta
\]
Let $z = x^2$ and replace $x$ by $\sqrt{z}$, we have:
\[
1 - \frac{1}{2Z_d} \int_{h^2}^{1} \frac{(1-z)^{(d-1)/2}}{\sqrt{z}} dz \le \delta^{1/m}
\]
Note that if we replace $\sqrt{z}$ by 1, LHS increases, so we can replace it with this stronger guarantee:
\[
1 - \frac{1}{2Z_d} \int_{h^2}^{1} (1-z)^{(d-1)/2} dz \le \delta^{1/m}
\]
Which becomes:
\[
\frac{1}{Z_d(d+1)} (1-h^2)^{(d + 1) /2} \ge 1 - \delta^{1/m}
\]
Note that: $1 - \delta^{1/m} = 1 - \exp(-(\log1/\delta) / m) < (\log1/\delta) / m$. So we can replace RHS with this stronger guarantee:
\[
(1-h^2)^{(d + 1) /2} \ge Z_d(d+1) (\log1/\delta) / m
\]
And $Z_d \le \eta/\sqrt{d}$ for some positive constant $\eta$ with sufficient large $d$, based on the two-sided Sterling formula. Plug this stronger guarantee into RHS:
\[
1 - h^2 \ge \big(\frac{\eta(d+1)}{\sqrt{d}}\frac{\log1/\delta}{m}\big)^{2/(d+1)}
\]
And with sufficiently large $d$, we can increase $\eta$ slightly to $\eta_1$ so that:
\[
h \le \sqrt{1 - \big(\frac{\eta_1 \sqrt{d} \log 1/\delta}{m} \big)^{\frac{2}{d + 1}}}
\]
To make RHS more comprehensible, we note that:
\begin{eqnarray*}
& 1 - \big(\frac{\eta_1 \sqrt{d} \log 1/\delta}{m} \big)^{\frac{2}{d + 1}} = & \\
& 1 - \exp(-2\frac{\log (m/\sqrt{d}) - \log(\eta_1\log 1/\delta)}{d + 1}) &
\end{eqnarray*}

For $0 \le x < 1$, we note that $1-e^{-x}$ is concave, and it is entirely above the line $(1-e^{-1})x$, i.e., $1-e^{-x} > (1-e^{-1})x$.
Plug this into RHS, we arrive at:
\[
h \le \sqrt{\alpha \max \big( \frac{\log (m / \sqrt{d})- \log(\eta_1 \log 1/\delta)}{d +1}, 0 \big)}
\]
Where $\alpha = 2(1-e^{-1})$.

\subsection{Proof of Theorem \ref{thm:random_centers_varying_norm}}

Fix an index $i$, we can divide the centers into two groups with norms $\{h_j\}_{1 \le j \le i}$ and $\{h_j\}_{(i+1) \le j \le m}$.
Let $j^* = \argmax_{1 \le j \le m} q \cdot c_j$, i.e, $j^*$ is the index of $c^*$, we have two cases:
\begin{itemize}
    \item $j^* \ge (i+1)$, i.e., the maximum inner product center is in the second group. We know that its inner product at least the largest among $i+1$ centers all with norms $\ge h_{j^*}$. This implies that $\cos(q, c^*) \ge L_1(i+1, \delta)$ with probability at least $1-\delta$.
    \item $j^* \le i$. Now the maximum inner product is in the first group.
    We generate a new set of centers with dividing every center in the first group by the smallest norm $h_i$, i.e., $\{ c_1 / h_i, c_2/h_i, \cdots, c_i/h_i \}$.
    Note that after division, all new centers still have norms $\ge 1$.
    The maximum inner product between the query and this new set of centers is at least as large as the maximum when all centers have unit norms.
    This implies that $q \cdot c^* / h_i \ge L_1(i, \delta)$ with probability at least $1-\delta$.
    Since $\|c^*\|_2 \le h_1$, we have $\cos(q, c^*) \ge \frac{h_i}{h_1} L_1(i, \delta)$ with probability at least $1- \delta$.
\end{itemize}
Combining these 2, we can conclude that $\cos(q, c^*) \ge \frac{h_i}{h_1} L_1(i, \delta)$ for any $i$ with probability at least $1-\delta$.
Hence the overall lower bound is $\max_{i \in [m]} \frac{h_i}{h_1} L_1(i, \delta)$.

\subsection{Proof of Theorem \ref{thm:optimal_direction}}

Without loss of generality, we can fix the norm of query at 1.
We can decompose the optimal direction $v = \alpha u + \beta w$, where $\alpha^2 + \beta^2 =1, \, \alpha, \beta >0$, and $w$ is a direction orthogonal to $u$.
We denote $A = \sqrt{\E_q ((q \cdot u)^2|c)}$ and $B=\sqrt{\E_q ((o^u_q \cdot w)^2|c)}$.
Then after some manipulation, we can arrive at:
\[
v^T \E_q(qq^T|c) v = \alpha^2 A^2 + \beta^2 B^2 + 2\alpha \beta\E_q ((q \cdot u)(o^u_q \cdot w)|c)
\]
Note that:
\[
\E_q ((q \cdot u)(o^u_q \cdot w)|c) \le \sqrt{\E_q ((q \cdot u)^2|c) \E_q ((o^u_q \cdot w)^2|c)} = AB
\]
So we have $v^T \E_q(qq^T|c) v \le (\alpha A + \beta B)^2$ even for the optimal $v$.
When we have $A >B$ or equivalently $A^2 > B^2$, we have $v^T \E_q(qq^T|c) v \le A^2$, but we also know that $v^T \E_q(qq^T|c) v \ge u^T \E_q(qq^T|c) u = A^2$.
Hence when $A^2 > B^2$, the optimal $v$ is equail to $u$ by setting $\alpha=1$.
And we have:
\begin{eqnarray*}
& B^2 = E_q ((o^u_q \cdot w)^2|c) = w^T \E_q (o^u_q(o^u_q)^T|c) w \le & \\
& \lambda_{\max}(\E_q (o^u_q(o^u_q)^T|c)) &
\end{eqnarray*}
So when the maximum eigenvalue of the matrix $\lambda_{\max}(\E_q (o^u_q(o^u_q)^T|c))  < \E_q((q \cdot u)^2|c)$, the optimal direction is equal to $u$.
Note that:
\begin{eqnarray*}
& \textrm{trace}(\E_q(o^u_q(o^u_q)^T|c)) = \textrm{trace}(\E_q(qq^T|c)) - & \\
& \textrm{trace}(\E_q((q \cdot u)^2 uu^T | c)) = 1 - \E_q((q \cdot u)^2|c) &
\end{eqnarray*}

And the smallest eigenvalue of matrix $\E_q(o^u_q(o^u_q)^T|c)$ is 0.
Hence it only has $(d-1)$ non-zero eigenvalues.
If the ratio between the largest and smallest non-zero eigenvalues of this matrix  is $\gamma$, then we have:
\[
\lambda_{\max} + (d-2)\lambda_{\max} / \gamma \le 1 - \E_q((q \cdot u)^2|c)
\]
Which gives us an upper bound of $\lambda_{\max}$ as:
\[
\lambda_{\max} \le \frac{1 - \E_q((q \cdot u)^2|c)}{1 + (d-2)/\gamma}
\]
So when RHS is less than $\E_q((q \cdot u)^2|c)$, we have the optimal direction is equal to the partition center direction.
After simplification, we arrive at the condition:
\[
\gamma < \frac{(d-2)\E_q((q \cdot u)^2|c)}{1 - 2\E_q((q \cdot u)^2|c)}
\]
We can replace RHS with a stronger gurantee and arrive at:
\[
\gamma < (d-2)\E_q((q \cdot u)^2|c)
\]
Now we can plug in the result from Theorem~\ref{thm:random_centers_varying_norm}, and conclude the proof.
\end{document}